\documentclass[sigconf]{acmart}

\usepackage{amsmath}
\usepackage{graphicx}
\usepackage{subcaption}
\usepackage{tabularray}
\usepackage{algorithm}
\usepackage{algpseudocode}
\usepackage{booktabs}
\usepackage{colortbl}
\usepackage{makecell}
\usepackage{float}
\usepackage{color,soul}
\usepackage{xcolor}
\usepackage{enumitem}

\setlist[itemize]{nosep,leftmargin=*,topsep=2pt,itemsep=1pt,parsep=0pt,partopsep=0pt}
\setlist[enumerate]{nosep,leftmargin=*,topsep=2pt,itemsep=1pt,parsep=0pt,partopsep=0pt}

\AtBeginDocument{%
  }

\copyrightyear{2026}
\acmYear{2026}
\setcopyright{cc}
\setcctype{by}
\acmConference[DAC '26]{63rd ACM/IEEE Design Automation Conference}{July 26--29, 2026}{Long Beach, CA, USA}
\acmBooktitle{63rd ACM/IEEE Design Automation Conference (DAC '26), July 26--29, 2026, Long Beach, CA, USA}
\acmDOI{10.1145/3770743.3804015}
\acmISBN{979-8-4007-2254-7/2026/07}

\begin{document}

\title{SPADE: An Input-Adaptive Sparse Attention Engine for Fast Video Diffusion Models Inference}

\author{Shanghao Liu\textsuperscript{1}, Renze Chen\textsuperscript{2}, Size Zheng\textsuperscript{2}, Yuanqiang Liu\textsuperscript{2}, Yun (Eric) Liang\textsuperscript{2}, Hailong Yang\textsuperscript{1}}
\authornote{Hailong Yang is the corresponding author.}
\affiliation{%
  \institution{\textsuperscript{1}Beihang University; \textsuperscript{2}Peking University\\
  Email: somehow6@buaa.edu.cn; \{crz, zhengsz, yuanqiangliu, ericlyun\}@pku.edu.cn; hailong.yang@buaa.edu.cn}
  \city{Beijing}
  \country{China}
}

\renewcommand{\shortauthors}{Liu et al.}

\begin{abstract}
Video diffusion transformers (vDiTs) generate high quality but pay quadratic self-attention cost, making inference prohibitive at video-token scales. The challenge is input-adaptive sparsity: selecting critical Q/K/V tokens with negligible overhead and executing them for end-to-end gains. We present \textsc{SPADE}, a training-free sparse-attention engine of three parts: (i) \textsc{vDiT-SSR}, spec defining 3D blocking candidates and formalizing dynamic masks via \emph{Summarizer}/\emph{Estimator} expressions; (ii) runtime \textsc{scheme generation} using SICS and a head-wise policy; and (iii) an executor with low-overhead index search, flash block-sparse attention, and kernel grouping. Across Hunyuan-Video, Wan~2.1/2.2 (T2V/I2V), \textsc{SPADE} raises sparsity and speed and preserves quality, accelerating attention $2.26{\times}$–$3.40{\times}$ and end-to-end $1.49{\times}$–$1.80{\times}$. Our code is open-sourced at \url{https://github.com/6somehow/DAC-SPADE}.
\end{abstract}



\maketitle

\section{Introduction}
Video generation has become a core generative-AI task. Modern systems—Sora~\cite{openai_sora}, Veo~\cite{Google_veo3}, Wan~\cite{wanteam_wan_2025}, Hunyuan-Video~\cite{kong_hunyuanvideo_2025}—typically use Video Diffusion Transformers (vDiTs) with self-attention.
Attention scales as $\mathcal{O}(n^2)$ in \emph{compute}; naive memory is also $\mathcal{O}(n^2)$, while streaming (FlashAttention) reduces memory to $\mathcal{O}(n)$ but leaves compute quadratic and the main latency source~\cite{dao_flashattention_2022}. In video DiTs, the 3D token grid aggravates both compute and traffic. Much of this cost is redundant~\cite{xiao_efficient_2024,jiang_minference_2024,deng_attention_2024}, motivating sparse attention for video~\cite{ding_efficient-vdit_2025,zhang_vsa_2025,tan_dsv_2025,chen_sparse-vdit_2025,yang_sparse_2025,xu_xattention_2025,xi_sparse_2025}.

We group sparse attention for video into: \textbf{i) Static}—fixed 3D blocking/sliding windows (e.g., Fast-VideoGen~\cite{zhang_fast_2025}); efficient but not adaptive. \textbf{ii) Semi-static}—pick among a few predefined patterns (e.g., Sparse-VideoGen~\cite{xi_sparse_2025}); mildly adaptive, but search space is tiny. \textbf{iii) Dynamic}—input-conditioned without a preset mask (e.g., Sparge Attention~\cite{zhang_spargeattn_2025}); flexible but with control/indexing overhead. The aim is higher sparsity and throughput without quality loss, yet most systems optimize one aspect in isolation. Progress toward near-linear scaling remains limited: sparsity gains are modest, and wall-clock speedups lag (Table~\ref{tab:motivation_results}).

\begin{table}
    \centering
    \small
    \setlength{\tabcolsep}{4pt}
    \setlength{\extrarowheight}{0pt}
    \addtolength{\extrarowheight}{\aboverulesep}
    \addtolength{\extrarowheight}{\belowrulesep}
    \setlength{\aboverulesep}{0pt}
    \setlength{\belowrulesep}{0pt}
    \caption{Comparison on Hunyuan-Video text-to-video (720p, 61~frames) among recent dense (FlashAttention-3), semi-static (Sparse-VideoGen), dynamic (Sparge Attention), and our \textsc{SPADE}.}
    \label{tab:motivation_results}
    \begin{tabular}{lccc} 
        \toprule
        Method                          & SSIM$\uparrow$ & Sparsity$\uparrow$ & Latency(ms)$\downarrow$  \\ 
        \midrule
        FlashAttention-3                & N/A            & 0\%                & 232             \\
        Sparse-VideoGen                 & 0.87           & 70.38\%            & 264             \\
        Sparge Attention                & 0.80           & 52.46\%            & 143             \\
        \rowcolor[rgb]{0.941,1,1} SPADE & 0.91           & 85.21\%            & 67              \\
        \bottomrule
    \end{tabular}
\end{table}
Combining \emph{static}, \emph{semi-static}, and \emph{dynamic} methods can raise sparsity but adds overhead, creating end-to-end challenges:
\textbf{(i) Blocking scheme selection.}
Choosing an input-conditioned 3D blocking from many candidates with negligible cost is nontrivial.
\textbf{(ii) Efficient dynamic kernels.}
Dynamic patterns improve adaptivity and sparsity but add control/indexing overhead, demanding hardware-aware kernels (accelerator instructions, fusion, access-aware design) and user customizability.
\textbf{(iii) Multi-granularity control.}
Token importance varies by head/layer/timestep~\cite{zhang_spargeattn_2025,li_snapkv_2024,cai_pyramidkv_2024,chen_sparse-vdit_2025}; exploiting this heterogeneity to increase sparsity while preserving quality remains hard.

We present \textsc{SPADE}, a training-free, algorithm--systems co-designed framework for high-performance hybrid-sparse attention. First, the \emph{Video DiT Sparsity Scheme Representation} (vDiT-SSR) is a unified interface spanning all three families, enabling flexible high-sparsity hybrids. vDiT-SSR expresses:
(i) diverse 3D static blocking via \emph{blocking-scheme candidates};
(ii) dynamic block-index selection via \emph{summarizer}/\emph{estimator} expressions; and
(iii) automatic sparsity control via \emph{policy-function} interfaces.

Built on vDiT-SSR, \textsc{SPADE} accelerates inference via two modules:
(1) \textsc{Scheme Generation}: fast, input-adaptive, head-wise/per-timestep patterns using the online Sum of Intra-block Cosine Similarities (SICS) with a lightweight \emph{policy} that budgets across timesteps/layers; and
(2) \textsc{Head-wise Sparse Attention}: an execution engine (codegen + library) that groups kernel launches, fuses QKV index-selection, and implements hardware-aware flash block-sparse attention for high-throughput adaptive computation.

We validate on T2V/I2V with Hunyuan-Video, Wan~2.1, and Wan~2.2. Compared to strong sparse baselines, \textsc{SPADE} preserves quality (VBench~\cite{huang2024vbench}, PSNR/SSIM/LPIPS) while achieving the highest sparsity and lowest latency. On an NVIDIA H800 at 720p, attention speedups are $2.26\times$--$3.40\times$ and end-to-end gains $1.48\times$--$1.70\times$.

In summary, our contributions are:
\begin{itemize}[nosep,leftmargin=5pt,itemindent=1pt,labelsep=0.5em]
\item \textbf{vDiT-SSR.} A unified, expressive module integrating 3D blocking candidates with dynamic selection (summarizer/estimator), subsuming prior designs and enabling new hybrids.
\item \textbf{Scheme Generation.} A head-wise, input-adaptive generator using online SICS plus a lightweight policy. Scheme generation is $8\%$ of sparse-attention runtime yet yields higher sparsity than alternatives.
\item \textbf{Head-wise Sparse Attention Execution.} Lowering vDiT-SSR to block-sparse kernels via codegen/library with low-cost index selection, kernel grouping, and hardware-aware optimizations. At equal sparsity, pattern search is $3.48\times$ faster and computation $1.80\times$ faster than state-of-the-art kernels.
\item \textbf{End-to-end gains.} On Hunyuan-Video and Wan~2.1/2.2 for T2V/I2V, \textsc{SPADE} speeds attention by $2.26\times$--$3.40\times$ and end-to-end by $1.49\times$--$1.80\times$, while matching or improving generation quality.
\end{itemize}

\section{Background}
\label{sec:background}

\subsection{Video Representation in vDiT Models}
\begin{figure}[thb]
	\centering
	\includegraphics[width=0.7\linewidth]{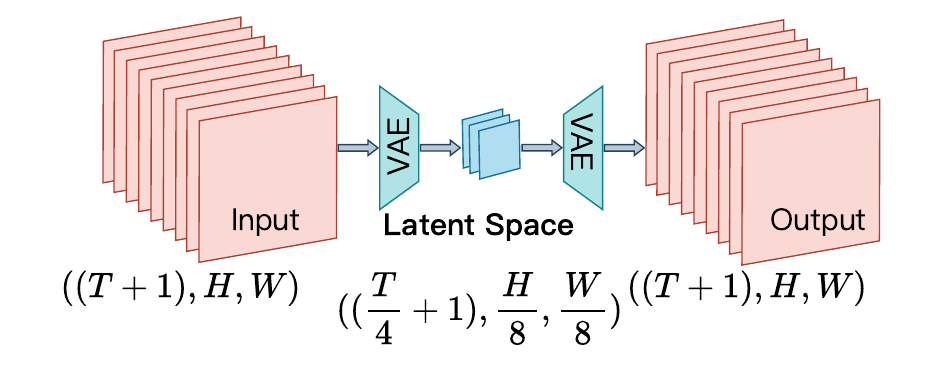}
	\caption{Illustration of the variational autoencoder (VAE) used in vDiT.}
	\label{fig:VAE}
\end{figure}
Variational Autoencoders (VAEs)~\cite{wu2018group} are integral to vDiTs, compressing raw video into compact latents for scalable training and inference. As in Figure~\ref{fig:VAE}, Hunyuan-Video~\cite{kong_hunyuanvideo_2025} and Wan~\cite{wanteam_wan_2025} map inputs to a 3D latent tensor $(T/4+1, H/8, W/8)$ that preserves spatio-temporal structure. The tensor is flattened to a 1D token sequence for attention. This preserved continuity is key to efficient sparse attention in video generation.

\subsection{Block-Sparse Attention}
The self-attention in Equation~\eqref{eq:attention} costs $\mathcal{O}(N^2)$ for sequence length $N$:
\begin{align}
	\text{Attention}(Q, K, V) = \operatorname{softmax}\left(\frac{QK^\top}{\sqrt{d_k}}\right)V \label{eq:attention}
\end{align}
Block-sparse attention (BSA) reduces this by partitioning tokens into blocks and masking a subset $M$ of block pairs (Equation~\eqref{eq:bsa}):
\begin{align}
	\text{Block-Sparse-Attn}(Q, K, V) = \operatorname{softmax}\left(\frac{QK^\top}{\sqrt{d_k}}\!+\!M \right)\!V \label{eq:bsa}
\end{align}
The complexity becomes $\mathcal{O}(NN_b)$, where $N_b$ is the number of tokens in selected blocks. Blockwise computation maps well to modern accelerators (e.g., NVIDIA tensor cores), sustaining high utilization while retaining long-range capacity.

\subsection{Taxonomy of Sparse Attention}
\label{sec:taxonomy_sparse_attention}

To target critical blocks in long-text and video, we group prior sparse attention into three classes.

\emph{Static methods} use input-agnostic masks (Attention Sink~\cite{xiao_efficient_2024}, DiTFastAttn~\cite{yuan2024ditfastattn}, Fast-VideoGen~\cite{zhang_fast_2025}). Fixed windows are simple and efficient but assume local stationarity and cannot adapt to content.

\begin{figure*}[t]
	\centering
	\includegraphics[width=0.93\textwidth]{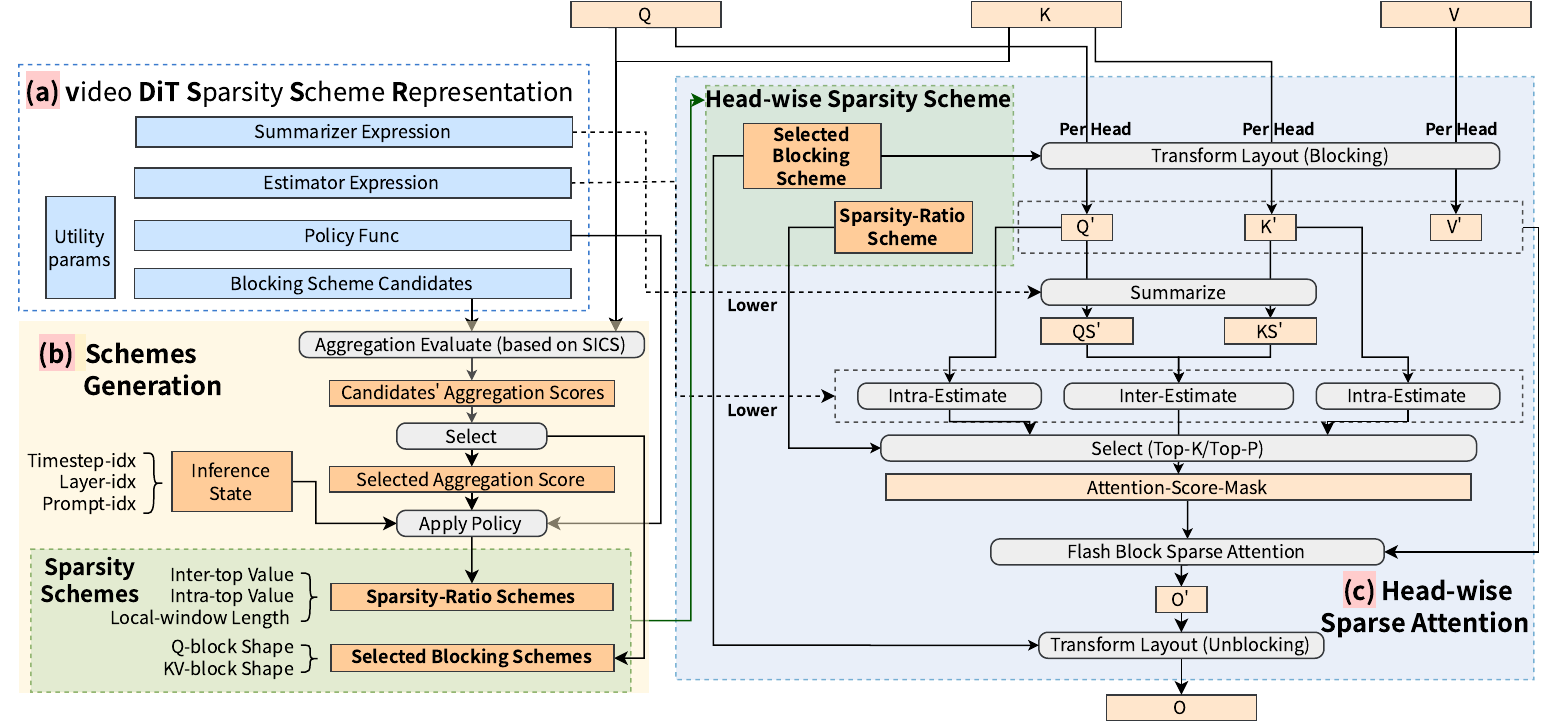}
	\caption{Overview of SPADE (Sparse Video DiT Engine).}
	\label{fig:method_overview}
\end{figure*}

\emph{Semi-static methods} choose among predefined masks per head using the input (DuoAttention~\cite{xiao_duoattention_2024}, DiTFastAttnV2~\cite{zhang2025ditfastattnv2}, Sparse-VideoGen~\cite{xi_sparse_2025}). They add limited adaptivity; mask selection can be inefficient or hurt accuracy.

\emph{Dynamic methods} generate masks from the input (ArkVale~\cite{chen_arkvale_nodate}, FlexPrefill~\cite{lai_flexprefill_2025}, SeerAttention~\cite{gao_seerattention_2025}, Sparge Attention~\cite{zhang_spargeattn_2025}). These often preserve quality with less compute, but mask construction adds overhead that degrades end-to-end latency.

\section{Methodology}

\begin{figure*}[t]
	\centering
	\begin{subfigure}[b]{0.285\textwidth}
		\includegraphics[width=\linewidth]{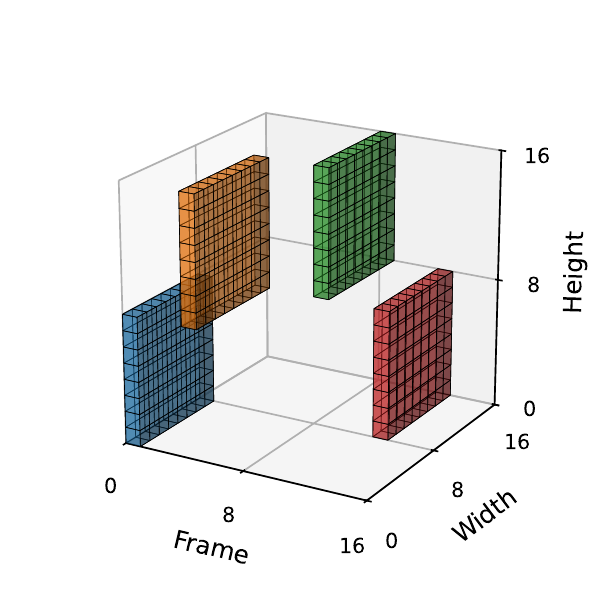}
		\caption{Example 0, Spatial scheme (1,8,8).}
		\label{fig:spatial_scheme}
	\end{subfigure}
	\hfill
	\begin{subfigure}[b]{0.285\textwidth}
		\includegraphics[width=\linewidth]{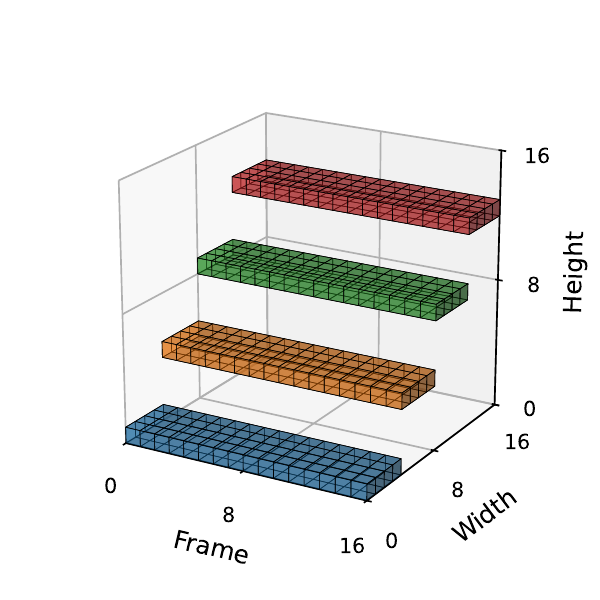}
		\caption{Example 1, Temporal scheme (16,1,4).}
		\label{fig:temporal_scheme}
	\end{subfigure}
	\hfill
	\begin{subfigure}[b]{0.285\textwidth}
		\includegraphics[width=\linewidth]{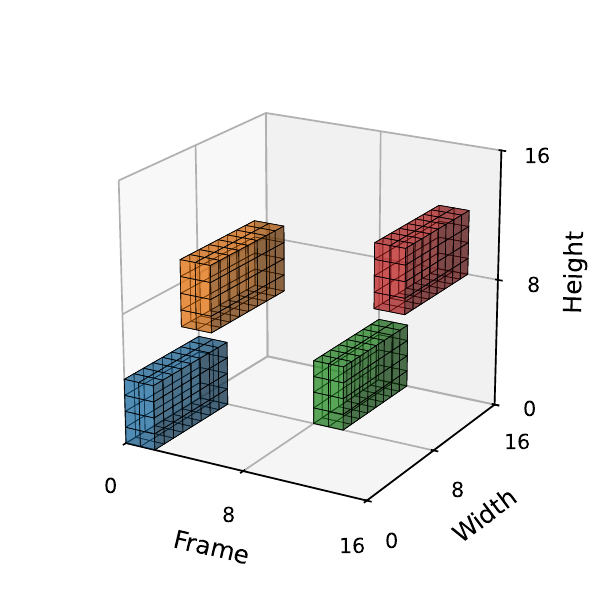}
		\caption{Example 2, Mixed scheme (2,4,8).}
		\label{fig:mixed_scheme}
	\end{subfigure}
	\caption{Illustration of blocking schemes (order: T,H,W): (a) spatial (1,8,8); (b) temporal (16,1,4); (c) mixed (2,4,8). Each color denotes one block.}
	\label{fig:blocking_schemes}
\end{figure*}

We introduce the \textbf{SPA}rse video \textbf{D}iT \textbf{E}ngine (SPADE), a sparse-attention engine for vDiTs (Figure~\ref{fig:method_overview}). SPADE has three components. Let the head dimension be $d_{head}$; Q, K, V the inputs and O the output; BS the block size and NB the number of blocks.

\textbf{(a) Video DiT Sparsity Scheme Representation (vDiT-SSR).} vDiT-SSR is an input spec for fine-grained sparsity. It lists \emph{blocking scheme candidates} and a \emph{policy function} for head-wise scheme generation, plus a \emph{Summarizer Expression} and an \emph{Estimator Expression} that formalize dynamic sparse attention.

\textbf{(b) Schemes Generation.} Given Q, K, and the inference state, this module performs \emph{Aggregation Evaluation} using the \emph{Sum of Intra-block Cosine Similarities} (SICS), then applies the policy to yield each head's \emph{sparsity-ratio scheme} and \emph{blocking scheme}.

\textbf{(c) Head-wise Sparse Attention.} This is the compute core: ahead-of-time compiled operators implementing the vDiT-SSR algorithm, including flash block-sparse attention and head-wise Top-K/Top-P selection. At runtime it consumes Q, K, V and the head-wise schemes to produce O.

\subsection{Video DiT Sparsity Scheme Representation (vDiT-SSR)}

Guided by Section~\ref{sec:taxonomy_sparse_attention}, vDiT-SSR unifies sparse attention for DiTs. As in Figure~\ref{fig:method_overview}, it specifies: (1) \emph{blocking scheme candidates} that partition Q/K into spatio-temporally continuous blocks; (2) a \emph{policy function} for static and dynamic sparsity; and (3) \emph{Summarizer}/\emph{Estimator} Expressions for dynamic methods. Utility parameters include attention-sink length~\cite{xiao_efficient_2024} and Top-K/Top-P mode~\cite{xu_xattention_2025}.

A \textbf{Blocking Scheme} partitions the latent spaces of Q, K, V into 3D blocks (Figure~\ref{fig:blocking_schemes}): \emph{spatial} (continuous in H,W; Figure~\ref{fig:spatial_scheme}), \emph{temporal} (continuous in T,W; Figure~\ref{fig:temporal_scheme}), or \emph{mixed} (spanning T,H,W; Figure~\ref{fig:mixed_scheme}). This flexible abstraction imposes minimal constraints.

Users enumerate \emph{blocking scheme candidates} for two reasons. Algorithmically, good partitioning benefits both static and dynamic selection by exploiting latent continuity. Abstraction-wise, it unifies static and semi-static patterns: e.g., Fast-VideoGen~\cite{zhang_fast_2025} equals a chosen scheme plus a local window; DiTFastAttn~\cite{yuan2024ditfastattn} and Sparse-VideoGen~\cite{xi_sparse_2025} are subsets of these schemes. Semi-static patterns arise from head-wise choice among candidates.

The \textbf{Policy Function} outputs head-wise \emph{sparsity-ratio schemes} from the \emph{Inference State} and the head’s \emph{Selected Aggregation Score}. The state includes \textit{timestep-idx}, \textit{prompt-idx} (user/negative), and \textit{layer-idx}. Each scheme has: (1) \emph{local-window length} (static); (2) \emph{intra-top value} (Top-P threshold or Top-K keep for dynamic intra); (3) \emph{inter-top value} (Top-P cumulative threshold or Top-K keep for inter).

The \textbf{Summarizer and Estimator Expressions} define the dynamic algorithm. The \emph{Summarizer Expression} reduces over sequence positions of Q/K (e.g., \textit{max}/\textit{min}/\textit{mean}). The \emph{Estimator Expression} includes intra-estimation on blocks (SICS, Eq.~\eqref{eq:SICS}) and a user-defined inter-estimator over Q/K summaries.

\begin{figure*}[t]
	\centering
	\begin{subfigure}[b]{0.285\textwidth}
		\includegraphics[width=\linewidth]{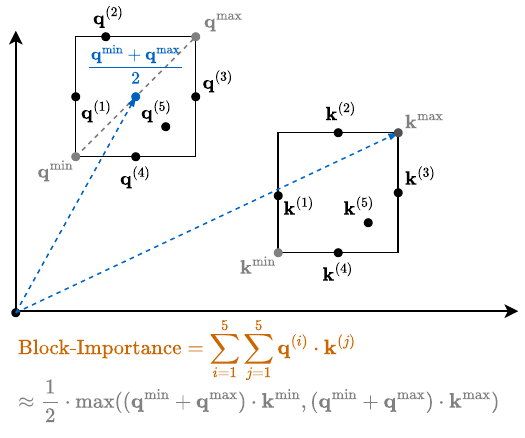}
		\caption{DSA-SPADE.}
		\label{fig:dsa-spade}
	\end{subfigure}
	\hfill
	\begin{subfigure}[b]{0.285\textwidth}
		\includegraphics[width=\linewidth]{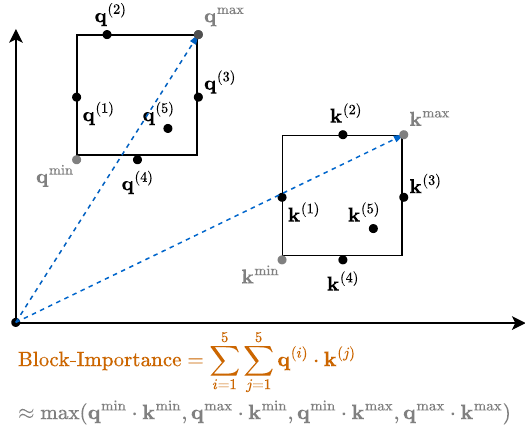}
		\caption{DSA2-SPADE.}
		\label{fig:dsa2-spade}
	\end{subfigure}
	\hfill
	\begin{subfigure}[b]{0.285\textwidth}
		\includegraphics[width=\linewidth]{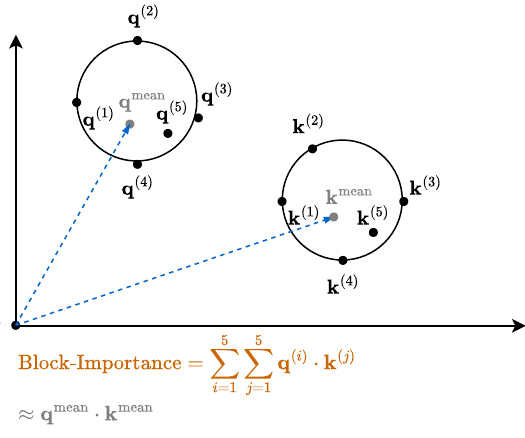}
		\caption{MEAN-SPADE.}
		\label{fig:mean-spade}
	\end{subfigure}
	\caption{Illustration of different summarizer and estimator expressions.}
	\label{fig:spade_expressions}
\end{figure*}

We denote variants as \emph{DSA-SPADE} (estimate with $(k_{\max},k_{\min} ,\tfrac{q_{\max}+q_{\min}}{2})$); \emph{DSA2-SPADE} (with $(k_{\max}, k_{\min}, q_{\max}, q_{\min})$); and \emph{MEAN-SPADE} (with $(q_{\mathrm{mean}}, k_{\mathrm{mean}})$). Figure~\ref{fig:spade_expressions} shows SPADE’s expressiveness: Figures~\ref{fig:dsa-spade},\ref{fig:dsa2-spade} are SPADE-inspired; Figure~\ref{fig:mean-spade} matches Sparge Attention~\cite{zhang_spargeattn_2025} and FlexPrefill~\cite{lai_flexprefill_2025}. Vectors lie in $\mathbb{R}^{d_{head}}$; black dots are per-block $\mathbf{q}^{(i)}, \mathbf{k}^{(i)}$; gray dots (e.g., $\mathbf{q}^{\text{max}}, \mathbf{k}^{\text{min}}$) are summaries. Gray equations indicate the inter-estimator; the dot product of blue vectors gives inter-block importance.

\subsection{Schemes Generation}
The module builds a per-head sparsity scheme. It evaluates all \emph{blocking scheme candidates} by aggregating Q and K under each, then picks the head-wise \emph{Selected Blocking Scheme} with the highest Aggregation Score (AS). For Q, AS is the sum of per-block SICS (Eq.~\eqref{eq:av_q}).
\begin{align}
    \mathrm{AS}\!\left(Q \,\middle|\, \text{scheme}\right)
    &= \sum_{i=1}^{NB} \mathrm{SICS}\!\left(\mathbf Q^{(i)}_{\text{block}}\right), \label{eq:av_q}\\[4pt]
    \mathrm{SICS}\!\left(\mathbf Q^{(i)}_{\text{block}}\right) &=  \mathbf 1_{BS}^{\!\top}\Big( \mathrm{CosSim}(\mathbf Q^{(i)}_{\text{block}},\mathbf Q^{(i)}_{\text{block}})\!\Big)\mathbf 1_{BS} \nonumber \\
    &= \mathbf 1_{BS}^{\!\top}\!\Big(D_i^{-1}\,\mathbf Q^{(i)}_{\text{block}}\big(\mathbf Q^{(i)}_{\text{block}}\big)^{\!\top} D_i^{-1}\Big)\mathbf 1_{BS}. \label{eq:SICS}
\end{align}
By Eq.~\eqref{eq:SICS}, let $\mathbf Q^{(i)}_{\text{block}}\in\mathbb R^{BS\times d_{\text{head}}}$, with $k$-th row $\big(\mathbf q^{(i)}_k\big)^\top$. Define $D_i=\operatorname{diag}\!\big(\lVert \mathbf q^{(i)}_1\rVert_2,\dots,\lVert \mathbf q^{(i)}_{BS}\rVert_2\big)$ and $\mathbf 1_{BS}\in\mathbb R^{BS}$. The operator $\mathrm{SICS}$ forms $D_i^{-1}\mathbf Q^{(i)}_{\text{block}}\big(\mathbf Q^{(i)}_{\text{block}}\big)^{\!\top}D_i^{-1}$ and sums all entries; $\mathrm{AS}(Q\,|\,\text{scheme})$ then sums over $NB$ blocks.

Intuitively, AS measures how tightly embeddings cluster within a spatio-temporal partition. Strong aggregation yields more representative summaries (better dynamic selection) and aligns static masks with latent structure. Thus head-wise scheme selection by aggregation benefits both static and dynamic sparsity.
\begin{align}
    \mathrm{SICS}\!\left(\mathbf Q^{(i)}\right)
    &= \sum_{a=1}^{BS}\sum_{b=1}^{BS}
       \frac{\mathbf q_a^\top \mathbf q_b}{\|\mathbf q_a\|_2\,\|\mathbf q_b\|_2} \nonumber\\
    &= \sum_{a=1}^{BS}\sum_{b=1}^{BS} \hat{\mathbf q}_a^\top \hat{\mathbf q}_b,
       \qquad \hat{\mathbf q}_k := \frac{\mathbf q_k}{\|\mathbf q_k\|_2} \nonumber\\
    &= \sum_{k=1}^{BS} \hat{\mathbf q}_k^\top \hat{\mathbf q}_k
       + 2\sum_{j=1}^{BS}\sum_{i=1}^{j-1} \hat{\mathbf q}_i^\top \hat{\mathbf q}_j \nonumber\\
    &= BS \;+\; 2 \sum_{j=1}^{BS} \Big(\sum_{i=1}^{j-1} \hat{\mathbf q}_i\Big)^\top \hat{\mathbf q}_j .
    \label{eq:SICS_optimize}
\end{align}
Naively, SICS costs $\mathcal{O}(BS^2 d_{head})$, which is expensive when screening many candidates. The \emph{online} form in Eq.~\eqref{eq:SICS_optimize} reduces this to $\mathcal{O}(BS d_{head})$ and lowers shared-memory pressure. $\mathrm{SICS}(\mathbf Q^{(i)})$ sums all pairwise cosine similarities; diagonal terms equal $BS$ since $\|\hat{\mathbf q}_k\|_2=1$. The final identity rewrites off-diagonals via cumulative partial sums for efficient computation.

As in Algorithm~\ref{alg:online_SICS}, our CUDA kernel uses implicit intra-warp sync, assigning each warp to a block and parallelizing across $d_{head}$. Float4 vectorization and warp shuffles improve throughput. Using registers only, it avoids shared memory and greatly cuts candidate-evaluation latency.
\begin{algorithm}[htb]
  \caption{Online computation of \(\mathrm{SICS}(\mathbf Q)\).}
  \label{alg:online_SICS}
  \begin{algorithmic}[1]
    \Require Block of vectors \(\mathbf Q=\{\mathbf q_0,\dots,\mathbf q_{BS-1}\}\subset\mathbb R^{d}\)
    \State \(\mathbf r \gets \mathbf 0 \in \mathbb R^{d}\) \Comment running partial sum of normalized vectors
    \State \(\sigma \gets 0 \in \mathbb R\) \Comment accumulator for off-diagonal inner products
    \For{$j \gets 0$ to $BS-1$}
      \State \(\hat{\mathbf q}_j \gets \mathbf q_j / \|\mathbf q_j\|_2\) \Comment normalize current vector
      \State \(\sigma \gets \sigma + \langle \mathbf r, \hat{\mathbf q}_j \rangle\)
      \State \(\mathbf r \gets \mathbf r + \hat{\mathbf q}_j\)
    \EndFor
    \State \textbf{return} \(2\,\sigma + BS\) \Comment equals \(\mathrm{SICS}(\mathbf Q)\)
  \end{algorithmic}
\end{algorithm}

After aggregation, SPADE selects the per-head \emph{Selected Blocking Scheme} (max AS). The \emph{Selected Aggregation Score} and the inference state then feed the policy function to produce each head’s \emph{sparsity-ratio scheme}, enabling fine-grained control over static vs. dynamic sparsity and overall levels.

\subsection{Head-wise Sparse Attention}
The performance core of SPADE is a highly optimized \textbf{Head-wise Sparse Attention} pipeline (Figure~\ref{fig:method_overview}c). Per head, Q, K, V are first rearranged into a blocked layout by the Selected Blocking Scheme for intra-block contiguity. The Summarizer Expression then produces one vector per Q/K block, yielding summaries $Q'$ and $K'$ of shape ($N_{blocks}$, $d_{head}$). From $Q'$/$K'$, we compute (i) \emph{intra-block importance} (one score per block) via the intra-estimator and (ii) an \emph{inter-block importance} matrix of size $N_{Q_{blocks}} \times N_{K_{blocks}}$ via the inter-estimator. Scores are filtered by Top-K or Top-P. SPADE supports arbitrary combinations of intra/inter importance; by default, intra scores are broadcast and \emph{OR}-combined with inter scores. The result is united with local-window and attention-sink masks to form a binary \emph{attention mask} defining the block-sparse pattern. A high-performance block-sparse attention kernel computes the main operation, after which output blocks are restored to the original layout to form $O$.

We apply three throughput optimizations. \textbf{(1) Operator fusion.} We fuse layout transform (blocking), summarization, and intra-estimation into one CUDA kernel, reducing global traffic, improving locality, and reusing on-chip memory. \textbf{(2) Head grouping.} At runtime, SPADE groups heads sharing a blocking scheme to increase per-launch work and utilization; this also lets us fuse Q/K gather–scatter with the layout transform, consolidating memory accesses. \textbf{(3) Flash block-sparse attention.} Inspired by FlashAttention~\cite{dao_flashattention_2022,flashattention2,flashattention3,spector2024thunderkittens} and sparse variants~\cite{wang2024flashmask,guo2024blocksparse}, our kernel (i) preprocesses the mask into ranked BSR for sequential access, (ii) uses an online $\operatorname{softmax}$ to avoid materializing dense attention, and (iii) exploits Hopper features (Tensor Memory Accelerator, warp-group MMA) to pipeline memory/compute. Heads with identical sparsity patterns are co-scheduled to raise overall utilization.


\section{Experiments}
\begin{table*}[ht]
    \centering
    \small
    \caption{Overall performance of sparse attention methods.}
    \label{tab:exp_overall}
    \begin{tabular}{l llllllll} 
      \toprule
      Method                                & SSIM $\uparrow$ & PSNR $\uparrow$ & LPIPS $\downarrow$ & \makecell{VBench\\score $\uparrow$} & \makecell{Attention\\TFLOPs $\uparrow$} & \makecell{Attention\\speedup $\uparrow$} & \makecell{E2E\\speedup $\uparrow$} & Sparsity $\uparrow$  \\ 
      \midrule
      \textbf{Hunyuan-T2V}                  &                 &                 &                    & 0.77                                                              & 704                                                                    & 1.00                                                                & 1.00                                                             & 0.00\%               \\
      SpargeAttn                            & 0.80            & 24.59           & 0.16               & 0.74                                                              & 246                                                                    & 0.88                                                                  & 0.80                                                             & 60.20\%              \\
      X-Attention                            & 0.88            & 27.38           & 0.09               & 0.76                                                              & 240                                                                    & 0.72                                                                  & 0.72                                                             & 52.46\%              \\
      Sparse-VideoGen                       & 0.87            & 26.26           & 0.11               & 0.76                                                              & 339                                                                    & 1.63                                                                  & 1.07                                                             & 70.38\%              \\
      Sparse-VideoGen2                      & 0.92            & 30.74           & 0.05               & 0.76                                                              & 248                                                                    & 2.17                                                                  & 1.23                                                             & 83.80\%              \\
      \rowcolor[rgb]{0.941,1,1} SPADE       & 0.91            & 29.04           & 0.08               & 0.76                                                              & 358                                                                    & 3.44                                                                  & 1.49                                                             & 85.21\%              \\
      \rowcolor[rgb]{0.941,1,1} SPADE-Turbo & 0.57            & 15.30           & 0.43               & 0.79                                                              & 358                                                                    & 3.44                                                                  & 1.80                                                             & 85.21\%              \\ 
      \midrule
      \textbf{Wan~2.1-T2V}                   &                 &                 &                    & 0.79                                                              & 696                                                                    & 1.00                                                                  & 1.00                                                             & 0.00\%               \\
      SpargeAttn                            & 0.76            & 22.59           & 0.23               & 0.72                                                              & 238                                                                    & 0.65                                                                  & 0.81                                                             & 47.10\%              \\
      Sparse-VideoGen                       & 0.73            & 19.97           & 0.25               & 0.79                                                              & 321                                                                    & 1.30                                                                  & 1.05                                                             & 64.46\%              \\
      Sparse-VideoGen2                      & 0.78            & 21.70           & 0.19               & 0.78                                                              & 195                                                                    & 1.57                                                                  & 1.11                                                             & 82.20\%              \\
      \rowcolor[rgb]{0.941,1,1} SPADE       & 0.87            & 25.87           & 0.10               & 0.79                                                              & 323                                                                    & 2.62                                                                  & 1.36                                                             & 82.32\%              \\
      \rowcolor[rgb]{0.941,1,1} SPADE-Turbo & 0.53            & 14.20           & 0.48               & 0.78                                                              & 323                                                                    & 2.62                                                                  & 1.49                                                             & 82.32\%              \\ 
      \midrule
      \textbf{Wan~2.2-T2V}                   &                 &                 &                    & 0.81                                                              & 701                                                                    & 1.00                                                                  & 1.00                                                             & 0.00\%               \\
      SpargeAttn                            & 0.66            & 18.72           & 0.35               & 0.78                                                              & 239                                                                    & 0.69                                                                  & 0.81                                                             & 50.60\%              \\
      Sparse-VideoGen                       & 0.47            & 12.62           & 0.54               & 0.81                                                              & 315                                                                    & 1.46                                                                  & 1.18                                                             & 69.18\%              \\
      \rowcolor[rgb]{0.941,1,1} SPADE       & 0.72            & 19.50           & 0.24               & 0.81                                                              & 302                                                                    & 2.56                                                                  & 1.32                                                             & 83.11\%              \\
      \rowcolor[rgb]{0.941,1,1} SPADE-Turbo & 0.48            & 12.81           & 0.56               & 0.81                                                              & 302                                                                    & 2.56                                                                  & 1.55                                                             & 83.11\%              \\ 
      \midrule
      \textbf{Wan~2.1-I2V}                   &                 &                 &                    & 0.71                                                              & 690                                                                    & 1.00                                                                  & 1.00                                                             & 0.00\%               \\
      SpargeAttn                            & 0.55            & 19.59           & 0.30               & 0.65                                                              & 229                                                                    & 0.90                                                                  & 1.17                                                             & 63.03\%              \\
      Sparse-VideoGen                       & 0.59            & 19.17           & 0.24               & 0.69                                                              & 321                                                                    & 1.31                                                                  & 1.22                                                             & 64.46\%              \\
      Sparse-VideoGen2                      & 0.64            & 20.34           & 0.20               & 0.70                                                              & 202                                                                    & 1.63                                                                  & 1.34                                                             & 82.00\%              \\
      \rowcolor[rgb]{0.941,1,1} SPADE       & 0.80            & 24.49           & 0.10               & 0.70                                                              & 276                                                                    & 2.26                                                                  & 1.41                                                             & 82.32\%              \\
      \rowcolor[rgb]{0.941,1,1} SPADE-Turbo & 0.53            & 17.99           & 0.26               & 0.69                                                              & 276                                                                    & 2.26                                                                  & 1.62                                                             & 82.32\%              \\ 
      \midrule
      \textbf{Wan~2.2-I2V}                   &                 &                 &                    & 0.73                                                              & 690                                                                    & 1.00                                                                  & 1.00                                                             & 0.00\%               \\
      SpargeAttn                            & 0.53            & 18.68           & 0.37               & 0.69                                                              & 232                                                                    & 0.89                                                                  & 0.88                                                             & 62.18\%              \\
      Sparse-VideoGen                       & 0.50            & 17.09           & 0.32               & 0.73                                                              & 313                                                                    & 1.28                                                                  & 1.21                                                             & 64.46\%              \\
      \rowcolor[rgb]{0.941,1,1} SPADE       & 0.79            & 24.66           & 0.10               & 0.73                                                              & 285                                                                    & 2.47                                                                  & 1.36                                                             & 83.25\%              \\
      \rowcolor[rgb]{0.941,1,1} SPADE-Turbo & 0.50            & 17.12           & 0.32               & 0.75                                                              & 285                                                                    & 2.47                                                                  & 1.49                                                             & 83.25\%              \\
      \bottomrule
    \end{tabular}
\end{table*}

\subsection{Experimental Setup}

\textbf{Models:}
We evaluate five open-source vDiTs: Wan~2.1-I2V/T2V-14B, Wan~2.2-I2V/T2V-14B, and Hunyuan-Video-T2V-13B. Videos are 720p. Wan~2.1/2.2 output 61 frames; Hunyuan-Video-T2V outputs 125. Each frame has 3600 tokens.

\textbf{Metrics:}
We report three dimensions:
\textbf{Sparsity}—fraction of pruned attention scores (theoretical speedup upper bound);
\textbf{Efficiency}—attention TFLOPs, attention-level speedup, and end-to-end speedup vs. full attention;
\textbf{Quality}—VBench~\cite{huang2024vbench} plus fidelity to full attention (SSIM/PSNR/LPIPS).

\textbf{Datasets:}
Text-to-Video uses VBench-2.0 prompts~\cite{zheng2025vbench} across 18 categories, with versions for Wan~2.1, Wan~2.2, and Hunyuan-Video. Image-to-Video uses VBench-2.0-I2V pairs~\cite{zheng2025vbench}, cropped to 16:9 for 720p.

\textbf{Configurations:}
All experiments run on an NVIDIA H800 (80GB, CUDA~12.8). Two fixed inference-time policies: \emph{Func~0} (Wan~2.1/2.2) budgets 30:1 for dynamic selection vs. static windows; \emph{Func~1} (Hunyuan-Video) budgets $\sim$80\% dynamic, 20\% static. Dynamic selection follows SPADE (Figure~\ref{fig:dsa-spade}): Summarizer and inter-estimator inputs; intra-block SICS; inter-block Top-K; intra-block thresholding.

\textbf{Baselines:}
We compare \textbf{Full Attention} via FlashAttention-3~\cite{flashattention3}; semi-static \textbf{Sparse-VideoGen}~\cite{xi_sparse_2025}; and dynamic \textbf{Sparge Attention}~\cite{zhang_spargeattn_2025}, \textbf{X-Attention}~\cite{xu_xattention_2025}, \textbf{Sparse-VideoGen2}~\cite{yang_sparse_2025}. X-Attention selects anti-diagonal (“slash”) scores; Sparse-VideoGen2 uses k-means to form the mask. Defaults are used. X-Attention is incompatible with Wan~2.1/2.2 and Sparse-VideoGen2 with Wan~2.2, so these are omitted. Following~\cite{yang_sparse_2025,xi_sparse_2025,li2024distrifusion,lv2024fastercache}, all methods use full attention for the first 30\% (T2V) and 25\% (I2V) denoising steps; \textit{SPADE-Turbo} sparsifies after the first step.

\subsection{Overall Performance}
\label{exp:overall}
Table~\ref{tab:exp_overall} summarizes results.

\noindent\textbf{Sparsity.}
\textit{SPADE} achieves the highest sparsity on every model (\textbf{82.32\%--85.21\%}), exceeding prior baselines (47.10--83.80\%). \textit{SPADE-Turbo} matches SPADE’s sparsity.

\noindent\textbf{Efficiency.}
Despite higher sparsity, SPADE’s kernels are most efficient: attention speedups \textbf{2.26$\times$--3.44$\times$} and end-to-end \textbf{1.32$\times$--1.49$\times$}, with the highest attention TFLOPs among dynamic baselines. \textit{SPADE-Turbo} further raises end-to-end speedup to \textbf{1.49$\times$--1.80$\times$} at the same sparsity.

\noindent\textbf{Quality (Overall + Fidelity).}
SPADE matches or surpasses baselines on perceptual quality: VBench is comparable to full attention (0.76--0.81); \textit{SPADE-Turbo} reaches 0.79 (Hunyuan-Video-T2V) and 0.75 (Wan~2.2-I2V). For fidelity, SPADE leads on Wan~2.1/2.2 (T2V/I2V) with higher SSIM/PSNR and lower LPIPS, and remains competitive on Hunyuan-Video-T2V (within \textbf{0.01} SSIM of the best sparse baseline).

\subsection{Ablation Study of vDiT-SSR Designs}
\label{exp:ablation}
\begin{table}[ht]
    \centering
    \small
    \setlength{\tabcolsep}{4pt}
    \caption{Ablation study of sparse attention methods of SPADE.}
    \label{tab:ablation}
    \setlength{\aboverulesep}{0pt}
    \setlength{\belowrulesep}{0pt}
    \begin{tabular}{l|lll}
    \toprule
    Method                              & SSIM & PSNR  & LPIPS  \\
    \hline\hline
    Static-Spatial                      & 0.56 & 17.08 & 0.46   \\
    Static-Temporal                     & 0.62 & 17.26 & 0.40   \\
    Static-Blocking                     & 0.63 & 18.34 & 0.35   \\
    Dync-Spatial                        & 0.81 & 23.55 & 0.15   \\
    Dync-Temporal                       & 0.81 & 23.70 & 0.16   \\
    \rowcolor[rgb]{0.941,1.0,1.0}SPADE  & 0.87 & 25.87 & 0.10   \\
    \bottomrule
    \end{tabular}
\end{table}

We ablate on Wan~2.1-T2V using SSIM/PSNR/LPIPS, comparing \textbf{Static-Spatial}, \textbf{Static-Temporal}, \textbf{Static-Blocking}, \textbf{Dync-Spatial}, \textbf{Dync-Temporal}, and \textbf{SPADE}. All maintain $\sim$82\% sparsity. Key findings: (1) Spatial/temporal variants validate the vDiT-SSR blocking candidates. (2) Static-Blocking vs. SPADE shows the benefit of SICS-based aggregation. (3) Dynamic variants beat static methods. (4) SPADE’s hybrid approach outperforms any single component.

\subsection{Attention Performance Breakdown}
\label{exp:breakdown}

\begin{figure}[ht]
\centering
\includegraphics[width=0.66\columnwidth]{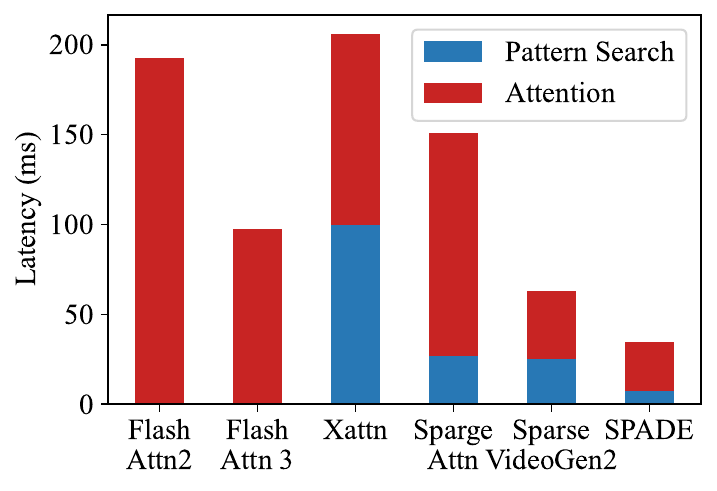}
\caption{Attention time breakdown of sparse attention methods.}
\label{fig:breakdown}
\end{figure}


We profile a single attention layer under the Wan~2.1 shape across FlashAttention-2/3, X-Attention, Sparge Attention, Sparse-VideoGen2, and SPADE. SPADE yields the lowest latency in \emph{Pattern Search} and \emph{Attention Computation}. Versus Sparse-VideoGen2, Pattern Search costs 25\% and Attention Computation 65\%; vs. Sparge Attention, Pattern Search is 26\%. These gains reflect SPADE’s algorithm–systems co-design. 



\subsection{Policy Function Analysis}
\label{exp:sensitivity}

\begin{figure}[ht]
\centering
\includegraphics[width=0.66\columnwidth]{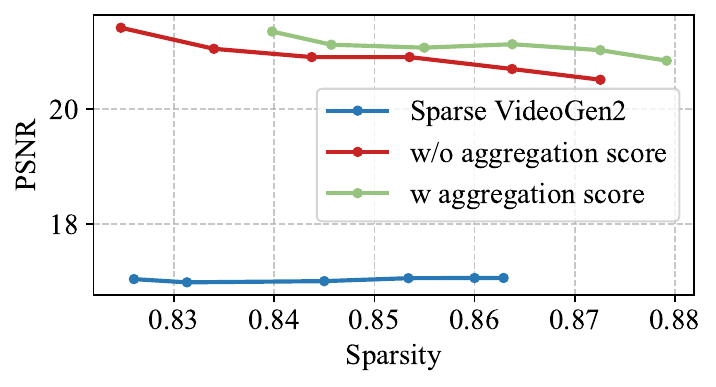}
\caption{Sensitivity test on Sparse-VideoGen2 and SPADE with policy functions with and without aggregation score.}
\label{fig:sensitivity}
\end{figure}

We study the policy that uses the Selected Aggregation Score to set head-wise Sparsity-Ratio Schemes. Using \textit{Complex\_plot} prompts from VBench-2.0 under high sparsity, we compare \textit{w/ aggregation score} (budget negatively correlated with the score) vs. \textit{w/o}, and include Sparse-VideoGen2. In complex scenes with tight budgets, aggregation-guided allocation yields higher fidelity and stability, validating SPADE’s policy design.


\section{Conclusion}

We introduced \textsc{vDiT-SSR}, a unified abstraction for composing \emph{static}, \emph{semi-static}, and \emph{dynamic} sparse attention in Diffusion Transformers. Built on it, \textsc{SPADE} pairs runtime \textit{Scheme Generation}—driven by SICS and a head-wise policy—with ahead-of-time \textit{Head-wise Sparse Attention} kernels, cleanly separating policy from mechanism and enabling efficient hybrid sparsity. 

Across Hunyuan-Video and Wan~2.1/2.2 on text-to-video and image-to-video, \textsc{SPADE} attains the highest sparsity and performance while matching or improving quality: attention accelerates $2.26\times$–$3.44\times$ and end-to-end $1.32\times$–$1.49\times$. \textit{SPADE-Turbo} further reaches $1.80\times$ end-to-end with a controlled quality trade-off.

\begin{acks}
This work is supported by National Natural Science Foundation of China (62322201 and U23B2020), the Fundamental Research Funds for the Central Universities (JKF-2025012343648 and JKF-20240598), and State Key Laboratory of Complex \& Critical Software Environment (SKLCCSE-2025ZX-04).
\end{acks}

\bibliographystyle{ACM-Reference-Format}
\bibliography{source/ref.bib}

\end{document}